# UTILISATION OF OPEN INTENT RECOGNITION MODELS FOR CUSTOMER SUPPORT INTENT DETECTION

Rasheed Mohammad, Oliver Favell, Shariq Shah, Emmett Cooper, Edlira Vakaj

Computer Science Dept., School of Computing and Digital Technology, Birmingham City University, Birmingham, UK

Rasheed.mohammad@bcu.ac.uk, oliver.favell@mail.bcu.ac.uk, shariq.shah@bcu.ac.uk, emmett.cooper@bcu.ac.uk, edlira.vakaj@bcu.ac.uk

## ABSTRACT

*Businesses have sought out new solutions to provide support and improve customer satisfaction as more products and services have become interconnected digitally. There is an inherent need for businesses to provide or outsource fast, efficient and knowledgeable support to remain competitive. Support solutions are also advancing with technologies, including use of social media, Artificial Intelligence (AI), Machine Learning (ML) and remote device connectivity to better support customers. Customer support operators are trained to utilise these technologies to provide better customer outreach and support for clients in remote areas. Interconnectivity of products and support systems provide businesses with potential international clients to expand their product market and business scale. This paper reports the possible AI applications in customer support, done in collaboration with the Knowledge Transfer Partnership (KTP) program between Birmingham City University and a company that handles customer service systems for businesses outsourcing customer support across a wide variety of business sectors. This study explored several approaches to accurately predict customers' intent using both labelled and unlabelled textual data. While some approaches showed promise in specific datasets, the search for a single, universally applicable approach continues. The development of separate pipelines for intent detection and discovery has led to improved accuracy rates in detecting known intents, while further work is required to improve the accuracy of intent discovery for unknown intents.*

## 1. INTRODUCTION

Customer support is a crucial need for businesses in the modern digital age where products are not just sold but also updated, repaired and maintained for their lifespan. Businesses have sought out new solutions to provide support and improve customer satisfaction as more products and services have become interconnected digitally. There is an inherent need for businesses to provide or outsource fast, efficient and knowledgeable support to remain competitive and maintain their consumer base in rapidly advancing and saturated consumer markets [1].
Support solutions are also advancing with technologies, including use of social media, Artificial Intelligence (AI), Machine Learning (ML) and remote device connectivity to better support customers. Customer support operators are trained to utilise these technologies to provide better customer outreach and support for clients in remote areas. Interconnectivity of products and support systems provide businesses with potential international clients to expand their product market and business scale.

As products become more advanced and require additional technical instruction, basic support requests are becoming automated to allow operators to focus on high priority and technically detailed requests. Automation systems aim to gain information from the customer and perform basic functionality such as altering account information to streamline the request process. AI could be leveraged to advance the functionality of support systems and help understand customer needs; this project aims to explore applications of AI within the customer support domain to improve and streamline the customer support process for greater customer satisfaction and efficiency.

Customer service technologies always strive to deliver faster, more efficient means of helping existing customers and facilitating new customers in order to stay competitive with other businesses. AI has found uses in automating previously manual sections of customer service roles often performed by humans, such as query identification and information acquisition, and in some cases automating entire services.

This paper reports the possible AI applications in customer support, done in collaboration with the Knowledge Transfer Partnership (KTP) program between Birmingham City University and a company that handles customer service systems for businesses outsourcing customer support across a wide variety of business sectors. The purpose is to help enhancing the usage of AI within their business and explore the possibilities of AI enhanced customer service solutions to benefit their main company processes.

The company offers bespoke customer service platforms, that if using an intent processing models could become more flexible to business requirements while improving request resolution speeds. This combined with potential cost saving from automated processes promotes their services to a wider market with potential for increased market growth and company revenue.

The business implications of an intent processing model would allow customer service providers to gather preliminary request information and narrow down the request type automatically, potentially automating simple requests and reducing waiting times for longer request types. Creation of a flexible model would facilitate optimisation of request processes across multiple business domains with minimal retraining, allowing more companies to automate their customer services using such a system. Such automation improves platform throughput, resulting in a faster customer service experience with higher service up time.

This project contributes to Deep Learning (DL) and Natural Language Processing (NLP) research in the area of open intent discovery, a domain focused on identification and extraction of contextual information from sentences often utilising deep networks and transformers. Researchers aim to improve the detection and extraction accuracy of unsupervised and semi-supervised models to eliminate dependency on models pre-trained with different contextual domain information. Using both semi-labelled and unlabelled data the model in this project explores prototype method combinations applied to the customer service requests domain, containing spoken dialogue and casual language semantics that few researchers have explored before.

This project focuses its research contribution towards evaluating the effectiveness of the intent classification method alongside potential performance improvements to the identification and extraction of intents. Any noted improvements could be applicable to the request's domain or to the wider intent discovery domain, marking significant progress in the fields of DL and NLP.

## 2. LITERATURE REVIEW

This research was conducted utilising [2]'s Preferred Reporting Items for Systematic Reviews and Meta-Analyses (PRISMA) method involving the collection of information resources, screening of eligible research, and analysis of both the qualitative and quantitative aspects of the resources (see Figure 1). Informal interviews were conducted with the company's employees and project associates, alongside regular progress meetings to outline business requirements regarding the project. Gathered requirements were factored into the aims and objectives to ensure the project achieves exploratory research outcomes and prototype implementation goals.

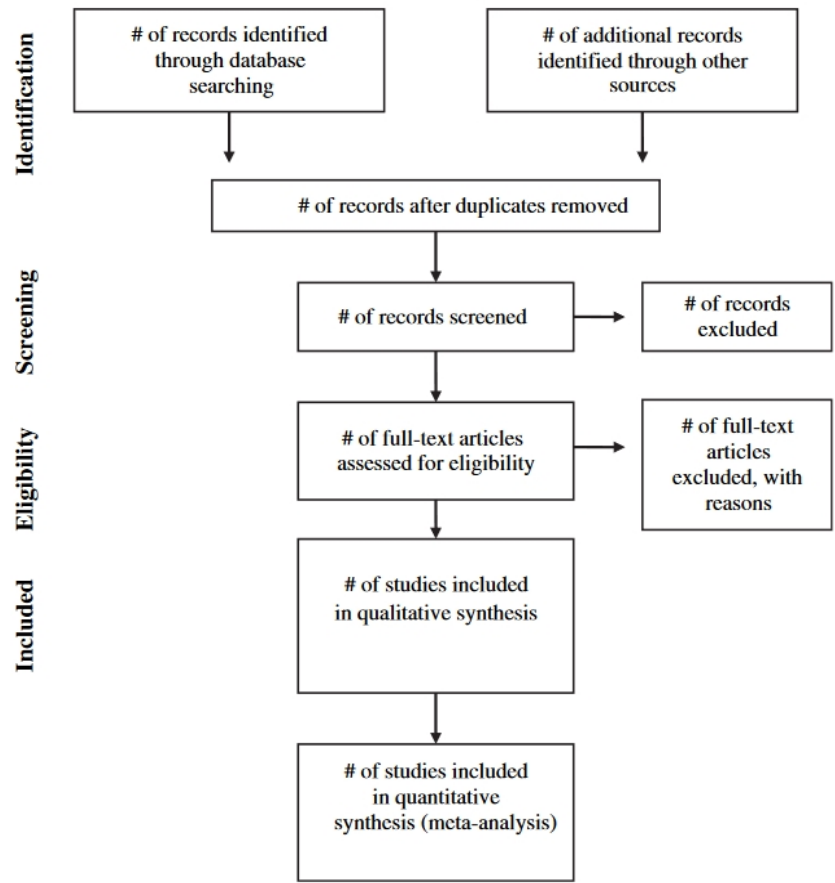

Figure 1: The PRISMA model, description and figure from [2]

## 2.1. Customer Support Systems (CSS)

The core principle that the business had to convey knowledge of the product's operation, maintenance and connectivity with other products led to a demand for fast, accessible, knowledgeable and reliable customer support operations [1, 3]. Originally CSS were operated by on-site experts, third-party resellers or software distributors providing support through phone lines during regular 9-5 business hours [1]; as technologies rapidly developed, support systems struggled to meet increased demand and provide customer contact points 24/7, impacting industries and consumers operating outside regular working hours [3]. Automated systems have been proposed to either fully or partially handle customer support requests to efficiency distribute informative product knowledge and expert help on-demand.

To provide an accessible, fast method of handling basic customer support requests like product information, customer account information or basic customer services, automated chat bots have been created to handle online messages through social media platforms and company support sites [3, 4]. Chat bots are available 24/7 and easily accessible even from mobile devices, providing customers with fast, reliable and remote support; functionality is however limited by internal AI programming, require domain knowledge to setup and often transfer complex requests to a human operator [3].

Other traditional methods of reducing reliance on live operators utilised online customer forums, message boards and facts and questions (FAQ) pages provide effective ways of customers finding answers to common questions to resolve product issues, alongside product reviews from real buyers [5]. Specialist knowledge and customer inquiries are initially required to build up a knowledge base of information, usually deployed within a company website or a regular technical support forum; this system creates an easily digestible and accessible customer contact point contrasting to technical specifications or lengthy support calls which some customers begrudge. Maintaining multiple contact points helps satisfy a wider array of customers and create well-rounded support systems capable of providing key information and upkeeping customer satisfaction.

## 2.2. Open Intent Recognition (OIR)

OIR is a new field in NLP focusing on the extraction and categorise of intentions from natural language statements using semi-supervised or unsupervised methods; this extends the intent analysis field which often requires expert domain knowledge and supervised training data for models to produce accurate results, creating inflexible models tailored to specific domains [6]. Extracted intents could then be used in dialogue systems to categorise statements [6, 7] and summarise large corpuses of natural language [8] without reliance on prior subject knowledge or supervised model training, pushing the boundaries of natural language understanding by machines.

There are three stages researchers identified in tackling this problem: intent detection, intent extraction and label classification. NLP models parse sentences to identify keywords, extract them in batches respective of their context, and label each set of keywords with a classification label such as "customer-book-hotel" indicating the intention. The parsed sentence, known as an utterance, can have several intents related to it depending on the complexity of the sentence or if it contains multiple subjects.

### 2.2.1. Intent Detection

To begin understanding natural language computers must first identify important words to contextualise the meaning of sentences. Sentences are passed through semantic parsers that evaluate each word through a series of gramatical trees with semantic rules to identify nouns, objects and verbs based on root forms; different grammatical parsers are used for differing grammatical rule structures such as Temizer and Diri [9]'s work in Turkish sentence parsing, and the Standford CoreNLP toolkit's annotator packages for various languages [10].

Parsers for utterance summarising aim to identify three key aspects which make up a semantic triplet: a subject (noun), a predicate (verb) and an object (statement) [11]. In contexts where the subject remains the same as in customer support, only the predicate and object need to be identified, so the resulting output is known as an action-object pair linked to the customer [6, 7]. Once all the key words in an utterance are parsed, triplets can be extracted using intent extraction methods.

### 2.2.2. Intent Extraction

Utterance pruning has to occur prior to extract, in which pronouns are resolved and linked to subjects to remove subject ambiguity and all verbs are converted to their root forms and similar meaning verbs are reduced using a VerbNet [8]. Extraction methods commonly iterate through utterances with grammatical rules to extract triplets, aiming to identify singular or multiple triplets based on identified subjects within the utterance. Ceran, et al. [8]'s work identified different events linked to each subject and extracted triplets using a semantic role labeller (SRL) and triplet matching rules, allowing for complex sentence analysis and multiple events contextually linked to multiple subjects. Rusu, et al. [11] used Treebank parsers which linked each word contextually to the subject, predicate or object in a sentence in a tree structure, and formatted triplets based on the resulting tree for each utterance, only extracting triplets from simple sentences with singular subjects.

### 2.2.3. Label Classification

Most discussed research focuses on identification and extraction of intents without any additional classification or categorisation processes. The work of Zhang, et al. [6], Liu, et al. [7] aims to tackle this, focusing on semi-supervised and unsupervised clustering of intents aiming to group extracted triplets into meaningful categories for human analysis or use in other systems. Liu, et al. [7] used unsupervised K-means clustering to group intents together, labelling each cluster with an action-object form label based on the representation percentage of top action-object pairs in the cluster; generated labels were evaluated against the human labelled ground truth to determine labelling accuracy.

Zhang, et al. [6] implemented a system with multiple clustering methods including Kmeans, hierarchical and density-based methods that utilised the KeyBERT toolkit [12] to label both utterances and clusters based on keyword representation; KeyBERT also provides a confidence score for the label based on the representation score of the label to the utterance or cluster. There are difficulties with both methods regarding the need for some labelled data to and difficulty handling large amounts of intent labels and distinguishing specific classes during classification [6].

### 2.2.4. Intent Recognition using Deep Learning

Both current state of the art models from Zhang, et al. [6], Liu, et al. [7] tackling the intent recognition problem proposed approaches utilising deep learning transformer frameworks; handling sentence parsing and keyword extraction based on the Bidirectional Encoder Representations from Transformers (BERT) framework [13]. Transformer frameworks are crucial as natural language is sequence and context based; transformers use an attention mechanism which "remembers" context from earlier sequence data when evaluating future information, retaining context helps understand the full meaning of utterances without segmenting them. Liu, et al. [7] leveraged the Siamese BERT (SBERT) framework [14] using Siamese neural networks to evaluate sentence embedding similarity and identify paraphrases to transform utterances. Semantic representations of utterances were classified using K-means clustering and clusters were labelled based on the most common unique identified action-object pairing within each cluster. Zhang, et al. [6] focused on a KeyBERT framework [12] to identify and extract key semantic words from utterances using a transformer neural network, alongside a K-means clustering approach similar to Liu, et al. [7]. Each cluster was aggregated to determine the top two key words within each cluster, which were joined together to form a two-part label for each utterance within that cluster. Both approaches leverage state of the art transformers and deep learning techniques to push the field of semantic intent analysis in a new direction of open intent recognition which seeks to develop a deeper understanding of sequential natural language data analytics.

## 3. RESEARCH METHODOLOGY

## 3.1. Architecture

The company wanted an online automated pipeline to process customer data effectively which also allowed company agents to access processed results to make decisions or provide customers with data insights. Clients would require a way to access the platform in addition to providing their data for processing, and in the long-term clients could access analytical insights post-processing. The models used would also be trained, stored and regularly retrained on newer training data all within the cloud so all model processing was computed in the cloud. The logical design is presented in figure 2.

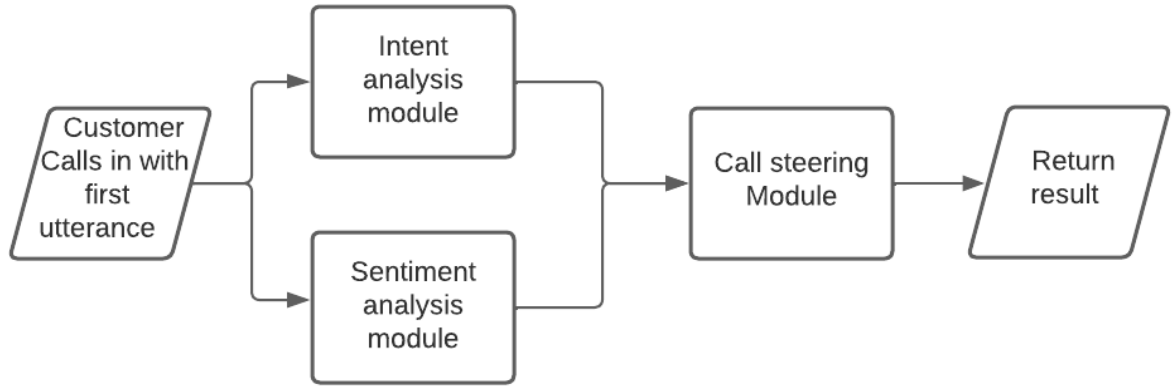

Figure 2: the logical flow of the project

The three main design architectures (figure 2) considered using company's internal servers as a system pipeline, utilising Google Cloud's VertexAI and Dialogflow systems for an online cloud-based data flow and a hybrid system involving Dialogflow and developing the TEXTOIR library [15] for deployment in an online container (figure 3). The most recommended approach focused on leveraging Google Cloud's Dialogflow alongside VertexAI to create an online system ready for data ingestion and capable of processing large quantities of data to form the groundwork of the rest of the project.

Though implementation designs differed all three approaches maintained the goals of providing a hosted platform for clients to connect to with different pipelines utilising client data for sentiment analysis, intent recognition and call routing. The costs of each of these services were broken down with pricing tables detailing service rates and the expected monthly cost at an estimated processing amount.

## 3.2. Data

Client data would be represented as raw audio files from call centres where an agent and customer are talking, these would be transcribed and then process by sentiment and intent analysis models. All data used would be stored within the cloud and only passed from one cloud service to another, with temporary processing data being discarded once model outputs were saved in post-processing (figure 3). In the long-term clients may also opt to archive their raw audio data on the cloud allowing full traceability between pre-processing and post-processed results; raw audio would be deleted after processing due to storage limitations as the transcription is still maintained alongside the results. The datasets employed in the experiments included ATIS [16], SNIPS [17], BANKING [18], HWU64 [19], CLINC150 [20], and real-life customer call data.

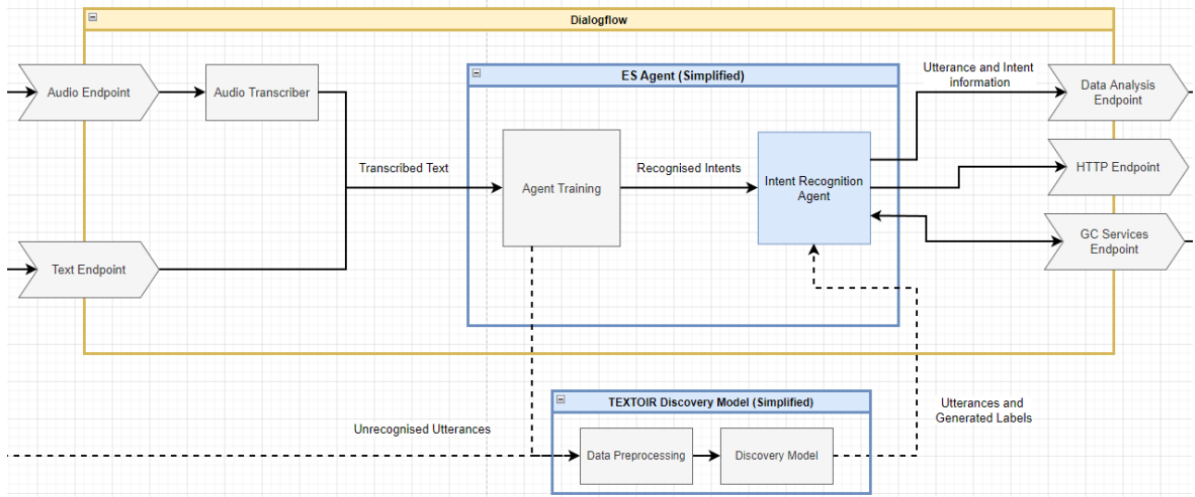

Figure 3:Dialogflow Pipeline using containerised TEXTOIR models

## 4. IMPLEMENTATION AND DISCUSSION

The KTP set out with the goal of making an online intent recognition platform providing a data pipeline that ingested raw audio data, converted it into text data before using machine learning models to recognize intents. The implementation of the KTP work produced an online platform on MS Azure, leveraging storage systems to store the raw data, transcribed text and model results all on the cloud. The machine learning model utilised in the processing containers was also uploaded to MS Azure, and processing was carried out using virtual machine instances, ensuring that the entire pipeline is hosted online. The pipeline enables clients to upload raw audio data to a container, which is then automatically processed and analysed, with the results saved into a MySQL database which is accessible via remote querying. The fundamental processing pipeline is therefore successfully set up and operating properly as per test runs conducted by the authors of this work and associates.

Initially, a supervised machine learning (ML) approach was taken, employing the transformers RoBERTa model [21] on a limited set of publicly available datasets. This approach achieved accuracy rates of 90-96%. However, since supervised ML requires labelled data, it was also necessary to experiment with unlabelled data, which poses a challenge for unsupervised ML. To address this challenge, another study was used [22], which advocated the use of unsupervised semantic clustering and dependency parsing. It employed several different combinations of pretrained models for semantic representation and clustering were explored, including Sentence-BERT, RoBERTa, Universal Sentence Encoder, K-means, Gaussian Mixture Model, and Hierarchical clustering. The approach was tested on six publicly available datasets and real-life data, performing well only on SNIPS and not on the other datasets, including real-life data. The unsupervised ML models employed in this study could only be slightly controlled using different techniques and parameter settings, and their performance was heavily dependent on the specific dataset used. Noisy data had a significant impact on their performance, and there was no single set of settings that could be used for all datasets. Consequently, the search for the best model continued.

Further research led to the creation of two separate pipelines for intent detection and discovery in PS5 as proposed in [6]. Their study involved used two different pipelines and experimented various semi-supervised or unsupervised ML algorithms. However, the best-performing algorithm, as reported in their study, was DA-ADB for intent detection and DeepAligned for intent discovery [23, 24]. The former was used to train models for detecting known and unknown intents, and the latter used for discovering unknown intents following intent discovery model training. Testing revealed that the DA-ADB model accurately detected the intents on which it had been trained, while classifying those on which it had not been trained as unknown. However, the intent discovery model did not perform well during training and evaluation, with longer training times and less accurate discovered intents than the intent detection model.

## 5. CONCLUSION

To address the problem of intent discovery, two techniques for intent generated labels were incorporated [25]: pattern for singularization task and WordNET (NLTK) for synonyms part. This was necessary, as generated intent labels were not identical for inputs that contained the same intent. The pattern for singularization achieved good performance, while WordNET did not perform well. The pattern successfully identified similarly-worded labels and considered them as one, but did not work on labels that were similar but positioned differently. However, processing times for these techniques were high without the use of CPU.

This study explored several approaches to accurately predict customers' intent using both labelled and unlabelled textual data. While some approaches showed promise in specific datasets, the search for a single, universally applicable approach continues. The development of separate pipelines for intent detection and discovery has led to improved accuracy rates in detecting known intents, while further work is required to improve the accuracy of intent discovery for unknown intents.

## AUTHORS

Rasheed mohammad: a computer science lecturer at Bcu, UK, specialized in NLP and data science.

Oliver Favell: a Master student at Bcu, UK and professional data science solution developer

Shariq Shah: Data Science specialist and AI-based solution developer

Emmett Cooper: an assistant lecturer at Bcu, A skilled Software Engineer with years of industrial experience

Edlira Vakaj: a Lecturer of Computer Science at the School of Computing and Digital Technology. She worked as a Marie Curie Experienced Researcher at the University of Surrey with secondment at Imperial College London in the area of Artificial Intelligence and Ontology Engineering. Edlira's current research interest is primarily in Artificial Intelligence, Semantic Web Technologies, and Knowledge Graphs.